\pdfoutput=1

\documentclass[11pt]{article}

\usepackage{acl}

\usepackage{times}
\usepackage{latexsym}
\usepackage{comment}
\usepackage{enumitem}
\usepackage{graphicx}
\usepackage{xspace}
\usepackage{caption}
\usepackage{subcaption}

\usepackage[T1]{fontenc}

\usepackage[utf8]{inputenc}

\usepackage{microtype}

\def\mystrut(#1,#2){\vrule height #1pt depth #2pt width 0pt}   

\definecolor{purple}{rgb}{0.5,0,1}
\definecolor{dcyan}{rgb}{0.2,0.6,0.5}
\definecolor{light-gray}{gray}{0.95} 

\definecolor{darkgreen}{RGB}{0,140,0}
\definecolor{darkred}{RGB}{200,0,0}
\definecolor{lightgreen}{RGB}{189,252,192}
\definecolor{lightred}{RGB}{255,205,212}
\definecolor{lightyellow}{RGB}{255,240,160}
\definecolor{lightblue}{RGB}{195,221,255}
\definecolor{lightpurple}{RGB}{232,209,255}

\newcommand{\redtext}[1]{\colorbox{lightred}{\mystrut(.5, .5) #1}}
\newcommand{\greentext}[1]{\colorbox{lightgreen}{\mystrut(.5, .5) #1}}

\newcommand{\bluetext}[1]{\colorbox{lightblue}{\mystrut(.5, .5) #1}}
\newcommand{\purpletext}[1]{\colorbox{lightpurple}{\mystrut(.5, 0.5) #1}}

\newcommand{\name}{\textsc{UnifiedQA}}
\newcommand{\namevtwo}{\textsc{UnifiedQA}-v2}

\title{\namevtwo: Better In/Out-of-Distribution Quality \\ via Broader Cross-Format Training}
\title{\namevtwo: \\ Better Generalization via Broader Cross-Format Training}
\title{\namevtwo: \\ Stronger Generalization via Broader Cross-Format Training}

\author{
  Daniel Khashabi$^{1}$ \hspace{0.4cm}
  Yeganeh Kordi$^{2}$ \hspace{0.4cm}   
  Hannaneh Hajishirzi$^{1,3}$ 
  \\
  \\
     $^1$Allen Institute for AI \hspace{0.5cm}  
     $^2$Tehran Polytechnic \hspace{0.5cm}
     $^3$University of Washington
}

\begin{document}

\maketitle
\begin{abstract}
We present \namevtwo{}, a QA model built with the same process as \name{}, except that it utilizes more supervision
-- roughly 3$\times$ the number of datasets used for \name. 
This generally leads to better in-domain and cross-domain results.\footnote{
\label{footnote:uqa:site}
\href{https://github.com/allenai/unifiedqa}{The models are accessible online.}
}
\end{abstract}

\section{Introduction}

An earlier work by~\citet{khashabi2020unifiedqa} showed transfer between different QA variants, such as extractive questions~\cite{rajpurkar-etal-2016-squad} and multiple-choice questions~\cite{lai-etal-2017-race}.
Based on this finding \citet{khashabi2020unifiedqa} presented \name, a single QA model trained on datasets of 4 different QA variants, and showed its empirical strength across a variety of benchmarks. 
Most notably, this approach showed remarkable generalization to unseen datasets compared to models specialized to individual datasets. 
Followup works have also replicated similar successes on more recent datasets that did \emph{not} exist at the time of \name's construction~\cite{hendrycks2020measuring,dasigi2021dataset,bragg2021flex,wu2021qaconv,zhong2021meta} -- further strengthening its empirical strength.

A question left open by~\citet{khashabi2020unifiedqa} 
is  whether there are gains in further broadening the supervision set; and if so, how sizable they are. 
We revisit this question by training a model (named \namevtwo) with the same process used in building \name{}, except that we use 20 datasets for training it (as opposed to 8 datasets used for supervising \name). 
Experimental results show that 
\namevtwo{} leads to 1-4\% performance improvements over \name{}, on average (columns indicated with `v2 - v1' in Table~\ref{tab:results}). 
The highest average gains correspond to mid-sized `large' models (4.2\% for in-domain and 4.5\% for out-of-domain setups

\section{Setup}
We construct a model using the same dataset encoding and hyperparameters as~\citet{khashabi2020unifiedqa}, except that we use larger supervision. 
Below we outline the relevant construction differences. 

The datasets user belong to one of the following four QA formats: \bluetext{extractive}, \redtext{abstractive}, \purpletext{multiple-choice} and \greentext{yes/no} questions. We henceforth follow this color-coding to indicate our perceived format of each dataset.

\subsection{Training the Models}
\label{subsec:training}
Like \name{}'s construction, we use the T5 architecture~\cite{raffel2019exploring} for \namevtwo{} and train it for 350k steps.\footnote{
While \name{} was trained for 120k steps, manual spot-checking on a few datasets indicated that \namevtwo{} requires more steps since it is trained a larger collection. 
} 
We do \emph{not} perform any dataset-specific model selection since the goal of the study is to have a \emph{single} model that works well across a variety of datasets.

\paragraph{Training datasets.}
\name{} is trained on the following 8 datasets:\footnote{The relevant citations are listed in Appendix~\ref{sec:appendix:datasets}. 
} 
\bluetext{SQuAD 1.1, SQuAD 2}, 
\redtext{NarrativeQA}, 
\purpletext{MCTest}, \purpletext{RACE, OpenBookQA, ARC}, 
\greentext{BoolQ}.\footnote{ARC and OpenBookQA datasets are used as-is without any retrieved evidence documents.}   



\noindent
\namevtwo{} is trained on following 20 datasets: 
\bluetext{SQuAD 1.1, SQuAD 2, NewsQA, Quoref, ROPES},  
\redtext{NarrativeQA, DROP, NaturalQuestions}, 
\purpletext{MCTest}, \purpletext{RACE, OpenBookQA, ARC, CommonsenseQA}, \purpletext{QASC, PhysicalIQA, SocialIQA, Winogrande},
\greentext{BoolQ, MultiRC (yes/no), BoolQ-NP}.


\begin{table*}[ht]
    \centering
    
    \begin{subfigure}[b]{\textwidth}
         \centering
         \vspace{-0.5cm}
         \caption{in-domain evaluation}
         \label{tab:in:domain}
         \vspace{-0.5cm}
        \includegraphics[scale=0.65, trim=1.71cm 14.25cm 1cm 1.4cm,clip=true]{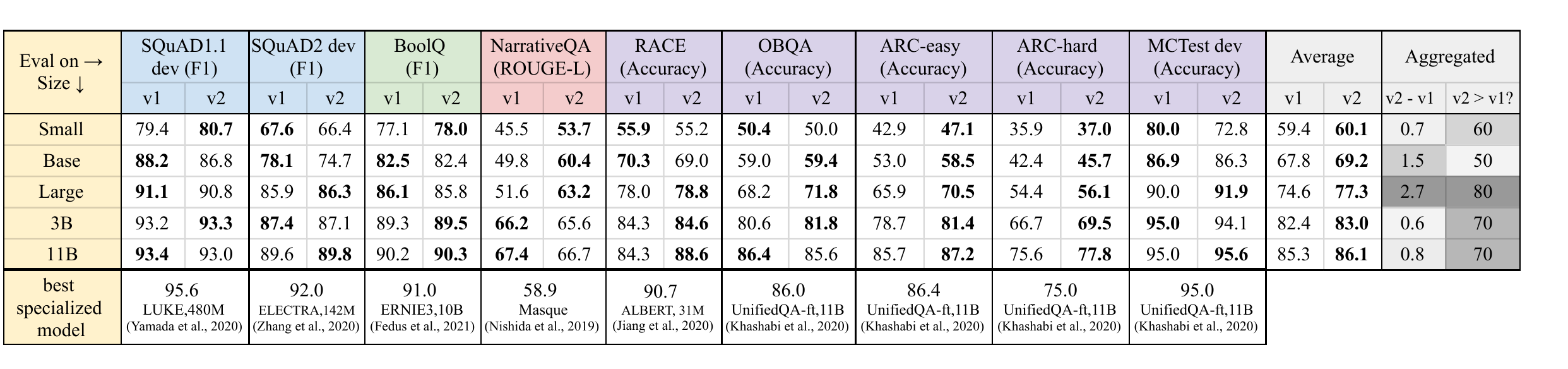}
        \vspace{-0.5cm}
    \end{subfigure}
     
    \begin{subfigure}[b]{\textwidth}
         \centering
         \caption{out-of-domain evaluation}
         \label{tab:out:domain}
         \vspace{-0.89cm}
        \includegraphics[scale=2.06, trim=1.77cm 18.03cm 18.34cm 1.53cm,clip=true]{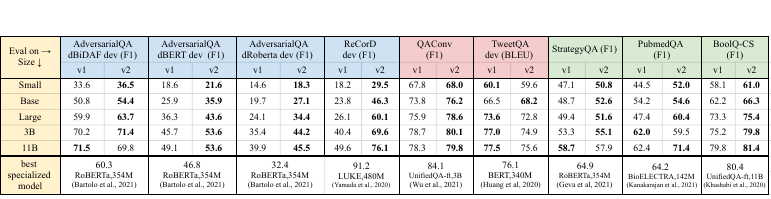}\\
         \vspace{-0.81cm}
        \includegraphics[scale=2.23, trim=10.48cm 18.03cm 10.23cm 1.53cm,clip=true]{figures/out-dist-table-2.pdf} \\ 
         \vspace{-0.81cm}
        \includegraphics[scale=2.125, trim=18.59cm 18.03cm 1.71cm 1.53cm,clip=true]{figures/out-dist-table-2.pdf}\\ 
        \vspace{-0.31cm}
    \end{subfigure}
    \caption{
        Evaluation of \name{} and \namevtwo{} models, 
        indicated with `v1' and `v2' headers, respectively. 
        \textbf{Neither of these models is specialized for individual datasets (i.e., one model evaluated across all datasets)}.
        The results are split across `in-domain' datasets  (observed during training) and 
        `out-of-domain' datasets  (\underline{not} observed during training).
        \namevtwo{} generally leads to better results, as indicated by the aggregate results in the last column. 
        For comparison, the bottom row indicates the best prior models that are often specialized (limited) to individual datasets.
        The colors indicate the formats: \bluetext{extractive}, \redtext{abstractive}, \purpletext{multiple-choice} and \greentext{yes/no}.
    }
    \label{tab:results}
\end{table*}



\subsection{Evaluation}
We discuss the evaluation setup for comparing \name{} and \namevtwo.
For each of these models, we evaluate a fixed checkpoint across all the target datasets: 
checkpoint 250k for \namevtwo{} and checkpoint 100k for  \name{},  following \citet{khashabi2020unifiedqa}'s setup for unseen dataset.


We evaluate the model for two settings:
(1) \emph{in-domain} evaluation on datasets seen by both models during training (discussed in \S\ref{subsec:training}). Summary of the results in Table~\ref{tab:in:domain}.
(2) \emph{out-of-domain} evaluation on datasets not seen by either of the models. In this setup: 
We evaluate the models on the following unseen 
\bluetext{AdversarialQA,
ReCoRD},
\purpletext{RACE-C}, 
\purpletext{
HeadQA, 
MMMLU, 
ReClor, 
Quail, 
OneStopQA,}
\purpletext{MCScript, 
MCScript 2.0,
CosmosQA,
DREAM,}
\purpletext{
ProcessBank, 
PROST,}
\greentext{
StrategyQA, 
PubmedQA,}
\redtext{
QAConv, 
TweetQA}.
The results are summarized in Table~\ref{tab:out:domain}. 
All numbers are reported on the test sets, except for those specified with `dev' which are evaluated on dev splits. 

\paragraph{Metrics.}
We evaluate each dataset via their common metric. The complete list of evaluation metrics is in Appendix~\ref{sec:appendix:metrics}. 
In addition to per-dataset scores, we provide aggregate scores that contrast the two models via a single number (the right-most columns of Table~\ref{tab:results}).
Specifically, we compute the difference between the average performance of v2 and v1 models of the same size (indicated with `v2 - v1') and 
the percentage of the datasets on which v2 leads to a better performance than v1 of the same size (indicated with `v2 > v1?'). 

\paragraph{Best prior specialized models.}
For ease of comparison, we include the best published results on each dataset (last rows in Table~\ref{tab:results}) which generally correspond to models that are fine-tuned to each dataset. 
While these are useful to understand the room for improvement on each dataset, they are not directly comparable to our target setup where we evaluate \emph{a single model across all datasets}. 

\nocite{yamada2020luke}

\nocite{zhang2020retrospective}

\nocite{sun2021ernie}

\nocite{nishida2019multistyle}

\nocite{jiang2020improving}

\nocite{Pan2019ImprovingQA}

\nocite{yamada2020luke}

\nocite{wu2021qaconv}

\nocite{huang2020nut}

\nocite{geva2021did}

\nocite{raj2021bioelectra}

\nocite{}

\nocite{pmlr-v101-liang19a}

\nocite{liu2020interpretable}

\nocite{rae2021scaling}

\nocite{}

\nocite{}

 \nocite{}

\nocite{}

\nocite{}

\nocite{}

\nocite{}

\nocite{}

\nocite{}

\section{Results}
\label{subsec:results}
We summarize our takeaways from Table~\ref{tab:results}. 
Across all experiments, \namevtwo{} leads to 1-4\% performance improvements over \name{}, on average (columns indicated with `v2 - v1'). 
The highest gains correspond to mid-sized `large' models (2.7\% for in-domain and 4.5\% for out-of-domain setups). Conversely, the lowest gains correspond to the extreme sizes (`small' and `11B'). 

Similar trends can also be seen with `v2 > v1' metric (percentage of datasets on which \namevtwo{} outperforms \name). On this metric, all the numbers are above 50\%, which indicates that v2 models  generally lead to higher performance. In particular, \namevtwo{} of size `large' outperforms \name{} of the same size on 80\% and 76\% of the datasets, for in-domain and out-domain settings, respectively.

The gains are not quite uniform across datasets. 
Among the in-domain datasets,  those with smaller supervision sets generally benefit more from broader training. For example, the `small' models on ARC-easy dataset (around 2k size) enjoy a gain of 4.2\% (42.9 $\rightarrow$ 47.1).
We leave further understanding of these gains as a function of various dataset properties to future work. 

\section{Related Work}
The past few years has seen a lot of activity toward  
\emph{unified} model designs~\cite{friedman2021single,aghajanyan2021muppet,gupta2021towards,tafjord2021general,lourie2021unicorn,jiang2021delphi,aribandi2021ext5,zhong2021adapting,chen2021meta,ruiqi2022difference}.  
For example, \citet{tafjord2021general} propose a unified model for simultaneously answering questions and explaining its decisions,   
\citet{zhong2021adapting} propose a zero-shot unified model for solving various classification tasks via an existing QA model, and 
\citet{aribandi2021ext5} show positive transfer across a wider variety of NLP tasks.


\section{Final Thoughts}
In this brief study, we showed the empirical benefits of broader training of QA systems, especially when evaluated for out-of-domain setup. The study leaves out several open questions for future work. 
One open question is whether broad and unified training of QA systems has any benefits in terms of compositional generalization~\cite{keysers2019measuring,gu2021beyond, bogin22generalization}, which is often pursued in the \emph{multi-hop} QA literature. 
We welcome more work in this direction. 

Another open direction is whether one can build unified models that can produce answers at different granularity~\cite{khashabi2021gooaq}. 
This necessitates models that accept more versatile inputs, ideally those that can encompass some definition of the expected outputs. 
This is reminiscent of the recent literature on solving tasks given their language instructions~\cite{mishra2021cross}. 
We hope to see variants of this literature tailored toward QA. 

Last but not least, 
it is not clear whether the gains here should be counted as 
meaningful progress toward models with better language understanding abilities
or 
mere memorizing of common questions that exploit the weaknesses of our (the research community's) evaluation protocols.\footnote{
``One man's generalization is another man's carefully engineered inductive bias'' --someone on Twitter
}
We hope that the future work addresses such important issues by  rethinking our means of measuring progress. 

\section*{Acknowledgment}
TPU machines for conducting experiments were provided by Google.

\bibliography{ref}
\bibliographystyle{acl_natbib}

\appendix

\section{Dataset References}
\label{sec:appendix:datasets}
For ease of readability in the main text, the citations corresponding to each dataset are included here: 
    SQuAD 1.1~\cite{rajpurkar-etal-2016-squad}, 
    SQuAD 2~\cite{rajpurkar-etal-2018-know}, 
    NewsQA~\cite{trischler-etal-2017-newsqa}
    Quoref~\cite{dasigi-etal-2019-quoref}, ROPES~\cite{lin-etal-2019-reasoning},
    AdversarialQA~\cite{bartolo2020beat},
    ReCoRD~\cite{zhang2018record},
    DROP~\cite{dua-etal-2019-drop}
    NarrativeQA/NarQA~\cite{kocisky-etal-2018-narrativeqa}, 
    QAConv~\cite{wu2021qaconv}, 
    TweetQA~\cite{xiong2019tweetqa}, 
    HeadQA~\cite{vilares2019head},
    RACE-C~\cite{pmlr-v101-liang19a},
    MCTest~\cite{richardson-etal-2013-mctest}, 
    RACE~\cite{lai-etal-2017-race}, OpenBookQA~\cite{mihaylov-etal-2018-suit}  ARC~\cite{clark2018think,clark2016combining},
    QASC~\cite{khot2019qasc},
    CommonsenseQA/CQA~\cite{talmor-etal-2019-commonsenseqa},
    Winogrande~\cite{sakaguchi2019winogrande},
    MMMLU~\cite{hendrycks2020measuring},
    ReClor~\cite{yu2019reclor},
    Quail~\cite{rogers2020getting},
    OneStopQA~\cite{starc2020},
    MCScript~\cite{ostermann2018mcscript},
    MCScript 2.0~\cite{ostermann2019mcscript2},
    CosmosQA~\cite{huang2019cosmos}, 
    ProcessBank~\cite{berant2014modeling}, 
    DREAM~\cite{sun2019dream}, 
    PROST~\cite{aroca2021prost}, 
    BoolQ~\cite{clark-etal-2019-boolq},
    BoolQ-NP~\cite{khashabi2020naturalperturbations}
    the binary (yes/no) subset of MultiRC~\cite{khashabi-etal-2018-looking},  
    StrategyQA~\cite{geva2021did}, 
    PubmedQA~\cite{jin2019pubmedqa}.

\section{Dataset Metrics}
\label{sec:appendix:metrics}
Here we summarize the metrics that were used for the evaluation.\footnote{Our evaluation script is public (see Footnote~\ref{footnote:uqa:site}).}
For each dataset, we have picked a metric that is recommended by dataset authors, and whenever possible, we use their script for evaluation. 
For multiple-choice tasks we always used accuracy: if the correct answer is picked, then the model gains one credit. Otherwise, no credit is given.
For extractive tasks, we use F1 token overlap between the answer text and gold answer(s). 
There are different ways of computing this score, usually based on how the texts are normalized. 
Therefore, for each dataset, we used its official F1 computation. 
For NarrativeQA we used ROUGE-L. 
For TweetQA we used BLEU. 
For Yes/No questions, we also use SQuAD 1.1's F1 token overlap: in most cases that the model provides ``yes''/``no'' answers, F1 acts like binary scoring (correct or incorrect credits). 
When the predictions contain more than ``yes''/``no'' (like, ``yes it is correct''), it provides a soft measure of correctness.

\end{document}